\documentclass{article}

\usepackage[a4paper,margin=3.5cm]{geometry} 
\usepackage{authblk}
\usepackage[english]{babel}
\usepackage{natbib}
\usepackage[utf8]{inputenc}
\usepackage[colorlinks=true,linkcolor=blue, citecolor=blue,backref=true]{hyperref}
\usepackage{csquotes}
\usepackage{lipsum}
\usepackage{amsmath,amssymb,amsthm}
\usepackage{enumitem}
\usepackage{graphicx}
\usepackage{algorithm}
\usepackage{algpseudocode}
\usepackage{subfig}
\algnewcommand\algorithmicinput{\textbf{Input:}}
\algnewcommand\Input{\item[\algorithmicinput]}
\algnewcommand{\LineComment}[1]{\State \(\triangleright\) #1}
\graphicspath{{./images/}}
\usepackage{mathtools}
\usepackage{xcolor}
\newtheorem{theorem}{Theorem}[section]
\newtheorem{definition}[theorem]{Definition}
\newtheorem{remark}[theorem]{Remark}
\newcommand{\indep}{\perp \!\!\! \perp}
\newcommand{\cG}{\mathcal{G}}
\newcommand{\cX}{\mathcal{X}}
\newcommand{\cP}{\mathcal{P}}
\newcommand{\cH}{\mathcal{H}}
\newcommand{\cZ}{\mathcal{Z}}
\newcommand{\cE}{\mathcal{E}}
\newcommand{\cD}{\mathcal{D}}

\title{A continuous Structural Intervention Distance to compare Causal Graphs}

\author[1]{Mihir Dhanakshirur}
\author[2]{Felix Laumann}
\author[3]{Junhyung Park}
\author[2]{Mauricio Barahona}
\affil[1]{Department of Mathematics, Indian Institute of Science, Bengaluru, India}
\affil[2]{Department of Mathematics, Imperial College London, London, UK}
\affil[3]{MPI for Intelligent Systems, T\"ubingen, Germany}

\date{}

\begin{document}

\maketitle

\begin{abstract}
    Understanding and adequately assessing the difference between a true and a learnt causal graphs is crucial for causal inference under interventions. 
    As an extension to the graph-based structural Hamming distance and structural intervention distance, we propose a novel continuous-measured metric that considers the underlying data in addition to the graph structure for its calculation of the difference between a true and a learnt causal graph. 
    The distance is based on embedding intervention distributions over each pair of nodes as conditional mean embeddings into reproducing kernel Hilbert spaces and estimating their difference by the maximum (conditional) mean discrepancy. 
    We show theoretical results which we validate with numerical experiments on synthetic data.
\end{abstract}

\section{Introduction}
In causal learning settings, we assume that data are generated according to a Structural Causal Model (SCM). 
The directional relationships between variables in an SCM originate from an underlying Directed acyclic graph (DAG) under the causal Markov assumption \citep[Section 6.5]{peters2017elements}. The data-generating DAG may thus be called the \emph{true} DAG. 
The task in any causal learning problem is to derive (or learn) this true DAG given access to the observational data generated by the underlying SCM. Hence, we call the result of the effort to derive the causal relationships embedded in the observational data the \emph{learnt} DAG.

In the present work, we are concerned with the problem of estimating the performance of a causal structure learning, or causal discovery algorithm by measuring its ability to accurately resemble the true DAG, including its potentially varying edge weights. 
Many widely used metrics exist \citep{peyrard2020ladder,acharya2018learning,singh2017comparative,garant2016evaluating,peters2015structural,acid2003searching}. 
However, the most prominent ones, the Structural Hamming Distance and the Structural Intervention Distance, are dominated by graph properties only and do not directly take the underlying data into account.  
The Structural Hamming Distance (SHD) is the square of the Frobenius norm  of the difference between the two (binary) adjacency matrices (of the true and learnt DAGs), i.e., it counts the number of edges in the learnt DAG that need to be added and removed so it is equal to the true DAG. On the other hand, the Structural Intervention Distance (SID) counts the number of pairwise interventional distributions on which the true DAG and the learnt DAG differ.

Our proposed distance, the \emph{continuous Structural Intervention Distance} (contSID), is based on both the graph and data properties by computing the distance between each pairwise interventional distribution implied by the observational distribution in the true and learnt DAGs.
The continuous SID has advantages over the SHD and SID, that are:
\begin{enumerate}
    \item Advantage over SHD: The goal of estimating a DAG from observational data is to later use it to estimate effects under interventions. However, the SHD merely calculates the number of changes in edges that are required to transform one DAG to another. Hence, two DAGs having the same SHD may still differ significantly in the interventional effects they imply. 
    \item Advantage over SID: The SID is computed based on a binary count (whether there is a difference in the effect or not) and cannot quantify the difference in interventional distributions inferred by the two DAGs---important when weights are expected to vary across edges.
    This poses a problem when practitioners are interested in the quantitative discrepancies between interventions. 
    The effect of an intervention beyond a binary count cannot be assessed without observational data, which we have access to because the original causal structure learning is conducted on observational data.
    
\end{enumerate}
\begin{figure}[!tbp]
  \centering
  \subfloat[$\cG_1$]{\includegraphics[width=0.3\textwidth]{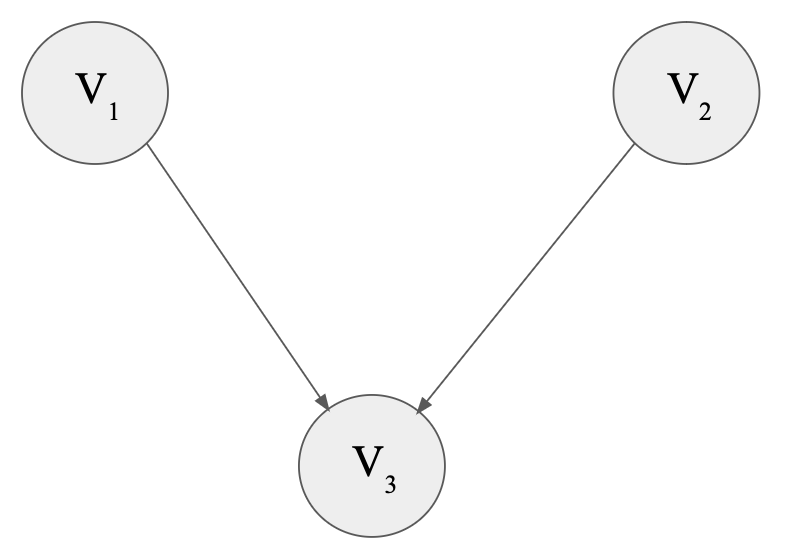}\label{fig: g1}}
  \hfill
  \subfloat[$\cG_2$]{\includegraphics[width=0.3\textwidth]{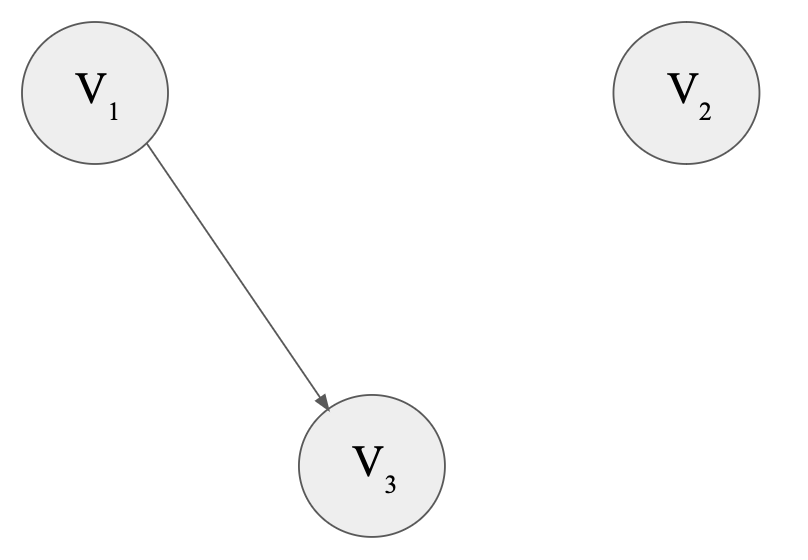}\label{fig: g2}}
  \hfill
  \subfloat[$\cG_3$]{\includegraphics[width=0.3\textwidth]{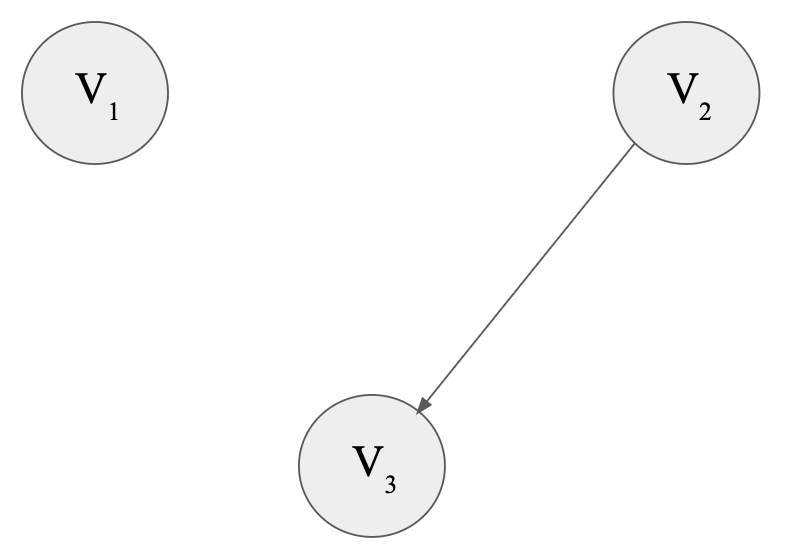}\label{fig: g3}}
  \caption{True DAG $\cG_1$ and learnt DAGs, $\cG_2$ and $\cG_3$ }
\end{figure}
\begin{table}[h!]
    \centering
    \begin{tabular}{l l  l}
        \hline
        Metric & $d(\cG_1,\cG_2)$ & $d(\cG_1,\cG_3)$ \\
        \hline
        SHD & 1 & 1  \\
        
        SID & 1 &  1 \\
        
        contSID & 0.23 & 0.39 \\
        \hline
    \end{tabular}
    \caption{SHD, SID and contSID calculated on $d(\cG_1,\cG_2)$ and $d(\cG_1,\cG_3)$.}
    \label{tab: comparison_table}
\end{table}
We demonstrate the issues of the SHD and SID by considering the following introductory example. We assume that data are synthetically generated by a linear model with additive Gaussian noise (\ref{eq: 1}) according to the DAG $\cG_1$ (Figure \ref{fig: g1}).
\begin{equation}\label{eq: 1}
    \begin{split}
        &V_1, V_2 \sim \mathcal{N}(0,1)\\
        V_3 &\sim \mathcal{N}(10V_1+V_2,1)
    \end{split}
\end{equation}
The edge connecting $V_1$ and $V_3$ has a mean ``weight'' of 10. 
Now, suppose $\cG_2$ (Figure \ref{fig: g2}) and $\cG_3$ (Figure \ref{fig: g3})  are two learnt DAGs (they could be the outcomes of two different causal discovery algorithms). 
We benchmark the quality of the learnt DAGs by comparing them across different metrics: Table $\ref{tab: comparison_table}$ describes the SHD, SID and contSID evaluated for the pair of DAGs $(\cG_1,\cG_2)$ and $(\cG_1,\cG_3)$. 
Intuitively, missing the edge $V_1 \to V_3$ should be penalized more than missing the edge $V_2 \to V_3$ since an intervention on $V_1$ would lead to a larger difference in the distribution of $V_3$ than the same intervention on $V_2$ (see Table~\ref{tab: comparison_table}). 
Hence, an appropriate metric should indicate that $\cG_2$ is a more accurate approximation of $\cG_1$ than $\cG_3$. However, both the SHD and the SID weigh missing the edges $V_1 \to V_3$ and $V_2 \to V_3$ equally. For a pair of DAGs, contSID quantifies the pairwise difference in the interventional distributions by using the observational distribution (via the valid adjustment set/backdoor set formula) as a mean embedding, that is, a unique representation of the interventional distribution in a reproducing kernel Hilbert space (RKHS).

As previously described in \citet{peters2015structural}, the SHD does not take into account the importance of the edge in terms of impact on the interventional distributions whereas the SID does.
However, the SID of $(\cG_1,\cG_2)$ and $(\cG_1,\cG_3)$ are still equivalent although missing the edge $V_1 \to V_3$ is clearly more influential on the resulting interventional distribution of $V_3$ than missing $V_2 \to V_3$.

We structure the paper as follows. After this Introduction, we provide sufficient Background in Section~\ref{background} to understand how we can use intervention mean embeddings (Section~\ref{IME}) to derive the Continuous Structural Intervention Distance in Section~\ref{contSID}. We demonstrate numerically the validity of our proposed metric (Section~\ref{experiments}) and conclude with a brief discussion (Section~\ref{Conclusion}).

\section{Background} \label{background}
We consider a finite collection of random variables $X_1, \dots, X_D$ with an index set $\mathbf{V} = \{1, \dots, D\}$. A graph $\cG = (\mathbf{V},\cE)$ then consists of nodes $\mathbf{V}$ and edges $\cE \subseteq \mathbf{V} \times \mathbf{V}$. We identify a node $V_j \in \mathbf{V}$ with its corresponding random variable $X_j$. We denote the parent set of a node $X_i$ by $\textbf{PA}_i\coloneqq\{X_j|(V_i,V_j)\in\cE,1\leq j\leq D\}$. We will use variables, nodes and vertices interchangeably depending on the context. We assume that the observational data $\mathcal{D} = \{x_1^{(n)},\dots,x_D^{(n)}\}_{n=1}^N$ are sampled from a distribution $P$ which has a density $p(\cdot)$ with respect to the Lebesgue or counting measure. Additionally, we require that the distribution is Markov with respect to the graph $\cG$.

\begin{definition}[Causal Markov assumption (\citet{peters2017elements}, Definition 6.21)] \label{def1}
    The distribution $P$ is Markov with respect to a DAG $\cG$ if $\textbf{A}\indep_{\cG}\textbf{B}|\textbf{C}\implies \textbf{A}\indep\textbf{B}|\textbf{C}$ for all disjoint vertex sets $\textbf{A},\textbf{B},\textbf{C}$, where $\indep_{\cG}$ denotes d-separation \citep[Definition 6.1]{peters2017elements}.
\end{definition}

The converse of the causal Markov assumption is known as the faithfulness assumption which links conditional independence in $P$ to d-separation in $\cG$. 
Both assumptions together imply the required intrinsic link between the existence of edges in a causal DAG and the joint distribution of the observed variables.

\begin{definition}[Faithfulness assumption (\citet{peters2017elements}, Definition 6.33)] \label{faithfulness}
    If two random variables are (conditionally) independent in the observed distribution $P$, then they are d-separated in the underlying DAG $\cG$.

\end{definition}

We also assume causal sufficiency, i.e., there are no hidden, or unobserved, variables that play a causal role in the system.

\subsection{Interventional distribution and \textit{do}-calculus}

Given random variables $X_i$ and $X_j$ where $i \neq j$, we try to estimate the distribution $P_{X_j|do(X_i) = \hat{x}_i}$, where $do(X_i) = \hat{x}_i$ represents an intervention on $X_i$ whose value is set to $\hat{x}_i$. This distribution is not directly observed since we are usually only given observational data. The \textit{do}-calculus \citep{pearl2009causality} enables us to estimate interventional distributions from observational distributions using a known DAG through valid adjustment sets \citep{peters2015structural}.

\begin{definition}[Valid adjustment set]\label{def0}
Let $X_j \notin \mathbf{PA}_i$ (otherwise we have $P_{X_j|do(X_i)} = P_{X_j}$, meaning interventions have no effect). We call a set $\mathbf{Z} \subseteq \mathbf{V} \setminus\{V_i,V_j\}$ a valid adjustment set for the ordered pair $(X_i,X_j)$ if 
\begin{equation}
    p(x_j|do(X_i) = \hat{x}_i)= \int_{\mathbf{z}} p(x_j|\hat{x}_i, \mathbf{z})p(\mathbf{z}). \label{eq: valid_adjust}
\end{equation}
\end{definition}
For discrete distributions, Equation~\eqref{eq: valid_adjust} becomes a summation instead of an integration. 
We can characterize valid adjustment sets using the following theorem.

\begin{theorem}[Characterization of valid adjustment sets \citep{peters2015structural,shpitser2012validity}]\label{th0}
    Consider a pair of variables $(X_i, X_j)$ and a subset $\mathbf{Z} \subseteq \mathbf{V}\setminus\{V_i,V_j\}$. Suppose $\mathbf{Z}$ satisfies the following property: In $\cG$, no $Z \in \mathbf{Z}$ is a descendant of any $X_k$ which lies on a directed path from $X_i$ to $X_j$(except for any descendants of $X_i$ that are not on a directed path from $X_i$ to $X_j$) and $\mathbf{Z}$ blocks all non-directed paths from $X_i$ to $X_j$.
    \newline 
    Then 
    \begin{itemize}
        \item If $\mathbf{Z}$ satisfies this property with respect to $(\cG, X_i, X_j)$, then $\mathbf{Z}$ is a valid adjustment set for $P_{X_j|do(X_i)}$.
        \item If $\mathbf{Z}$ does not satisfy this property with respect to $(\cG, X_i, X_j)$, then there exists a distribution $P'$ (not necessarily equal to $P$), with density $p'$, that is Markov with respect to $\cG$ and leads to $p'(x_j|do(X_i = \hat{x}_i) \neq \int_{\mathbf{z}} p'(x_j|x_i,\mathbf{z})p'(\mathbf{z})$, i.e., $\mathbf{Z}$ is not a valid adjustment set.
    \end{itemize}
\end{theorem}
Note that for a pair of nodes $(X_i, X_j)$ there exist many valid adjustment sets. The parent adjustment set, formed by taking $\mathbf{Z}$ to be the set of parents $\mathbf{PA}_i$ of $X_i$ is a valid adjustment set that can be easily read off from a graph.

\subsection{Conditional mean embeddings and the MCMD}

A mean embedding is a mapping of a probability distribution into an RKHS by a kernel k. This mapping is one-to-one if the kernel is characteristic \citep{fukumizu2007kernel}. 
 We adopt the measure-theoretic approach to kernel conditional mean embeddings \citep{park2020measure}, rather than the definition based on operators between RKHSs as introduced by \citep{song2009hilbert}. 
The measure-theoretic approach has the advantage of not relying on stringent assumptions for the population version of the embedding to exist, and comes with a natural regression interpretation for empirical estimates. 
\par
The maximum (conditional) mean discrepancy (MMD) is a measure of discrepancy between distributions that is widely-used in the machine learning community due to its elegance, attractive theoretical properties and ease of empirical estimation, and forms the backbone of our approach in this paper; however, we do note that there are many other measures of discrepancy between distributions, and leave it as interesting future research direction to investigate how those can be utilised for the problem we tackle in this paper.
In this section, we present the preliminaries of the conditional mean embedding and discuss its empirical estimates in Section~\ref{empirical-estimates}. The results presented here hold generally---we adapt them to our setting in Section~\ref{IME}.

As in \citet{park2020measure}, let $(\Omega,\mathcal{F},\cP)$ be the underlying probability space, let $(\cX,\mathfrak{X})$ and $(\cZ,\mathfrak{Z})$ be separable measurable spaces, and let $X: \Omega \rightarrow \cX$ and $Z: \Omega \rightarrow \cZ$ be random variables with distributions $P_X$ and $P_Z$. Let $\cH_{\cX}$ be a vector space of $\cX \rightarrow \mathbb{R}$ functions endowed with a Hilbert space structure via an inner product $\langle \cdot,\cdot \rangle_{\cH_{\cX}}$. A symmetric function $k_{\cX}: \cX \times \cX \rightarrow \mathbb{R}$ is a reproducing kernel of $\cH_{\cX}$ if and only if (i) $\forall x \in \cX, k_{\cX}(x,\cdot) \in \cH_{\cX}$; and (ii) $\forall x \in \cX \text{ and } \forall f \in \cH_{\cX}, f(x) = \langle f,k_{\cX}(x,\cdot) \rangle_{\cH_{\cX}}$. 

\begin{definition}[Kernel mean embedding] Given a distribution $P_X$ on $\cX$ and assuming $\mathbb{E}_{X}[k_{\cX}(X,X)]<\infty$, we define the kernel mean embedding of $P_X$ as $\mu_{P_X}(\cdot) = \mathbb{E}_X[k_{\cX}(X,\cdot)]$   
\end{definition}
\begin{definition}[Characteristic kernel] A positive definite kernel $k_{\cX}$ is characteristic to a set $\cP$ of probability measures on $\cH_{\cX}$ if the map $\cP \to \cH_{\cX}: P_X\mapsto \mu_{P_X}$ is injective.
\end{definition}
Popular kernels like the Gaussian and Laplacian kernel are characteristic.
The RKHS associated with a characteristic kernel is rich enough to enable us to distinguish between different distributions using their embeddings. In other words, we can define the MMD, on $\cP$: for $P_X,P_{X'} \in \cP$, let $||\mu_{P_X}-\mu_{P_{X'}}||$ be their MMD. 

\begin{definition}[Conditional mean embedding \citep{park2020measure}]
    Suppose $X$ satisfies $\mathbb{E}_X[k_{\cX}(X,X)] < \infty$. Then, we define the conditional mean embedding of $X$ given $Z$ as:
    \begin{equation}
        \mu_{P_{X|Z}} \coloneqq \mathbb{E}_{X|Z}\left[k_{\cX}(X,\cdot)|Z\right] \label{eq2}
    \end{equation}
\end{definition}

The conditional mean embedding $\mu_{P_{X|Z}}$ is a $Z$-measurable random variable taking values in $\cH_{\cX}$. The following theorem is used in estimating the conditional mean embedding (CME) of the conditional distribution $P_{X|Z}$.

\begin{theorem}[Deterministic function of conditional mean embedding \citep{park2020measure}]\label{th1}
    Denote the Borel $\sigma$-algebra of $\cH_{\cX}$ by $\mathcal{B}(\cH_{\cX})$. Then we can write $\mu_{P_{X|Z}} = F_{P_{X|Z}} \circ Z$, where $F_{P_{X|Z}}:\cZ \rightarrow \cH_{\cX}$ is some deterministic function, measurable with respect to $\mathfrak{Z}$ and $\mathcal{B}(\cH_{\cX})$.
\end{theorem}

For $z\in\cZ$, $F_{P_{X|Z}}(z) = \mathbb{E}_X[k_{\cX}(X,\cdot)|Z=z] = \mu_{P_{X|Z=z}}$ which is the kernel mean embedding of the distribution $P_{X|Z=z}$. Consider the random variables $X':\Omega\to\cX$ and $Z':\Omega\to\cZ$ with $E_{X'}[k_{\cX}(X',X')]<\infty$. By Theorem~\ref{th1}, $\mu_{P_{X'|Z'}} = F_{P_{X'|Z'}}\circ Z'$. The analog to the MMD for conditional distributions $P_{X|Z}$ and $P_{X'|Z'}$, the maximum conditional mean discrepancy (MCMD), is defined below:

\begin{definition}[Maximum conditional mean discrepancy \citep{park2020measure}]
The maximum conditional mean discrepancy (MCMD) between $P_{X|Z}$ and $P_{X'|Z'}$ is the function from $\cZ\to\mathbb{R}$ defined by
\begin{equation}
    \text{MCMD}_{P_{X|Z},P_{X'|Z'}}(z)= ||F_{P_{X|Z}}(z)-F_{P_{X'|Z'}}(z)||_{\cH_{\cX}}
\end{equation}
\end{definition}
Note that the MCMD at $z\in\cZ$ is equal to the MMD between the distributions $P_{X|Z=z}$ and $P_{X'|Z'=z}$. We use this later in section \ref{empirical-estimates} to construct a plug-in estimate of the MMD.

\subsection{Empirical estimates}\label{empirical-estimates}
By Theorem~\ref{th1}, the task of estimating $\mu_{P_{X|Z}}$ has been simplified to estimating $F_{P_{X|Z}}:\cX \to \cH_{\cX}$. 
This is precisely the setting of vector-valued regression with input space $\cX$ and output space $\cH_{\cX}$. The problem of estimating $F_{P_{X|Z}}$ can be reformulated as finding the vector-valued function that minimizes the loss $\cE_{X|Z}(F) \coloneqq E_Z\left[||F_{P_{X|Z}}(Z) - F(Z)||_{\cH_\cX}^2\right]$ among all $F \in \cG_{\cX\cZ}$, where $\cG_{\cX\cZ}$ is a vector-valued RKHS of functions $\cZ \rightarrow \cH_{\cX}$. 
For simplicity, we endow $\cG_{\cX\cZ}$ with a kernel $l_{\cX\cZ}(z,z') = k_{\cZ}(z,z') \; I'$ where $k_{\cZ}(\cdot,\cdot)$ is a scalar kernel on $\cZ$ and $I'$ is the identity operator.
\par
We cannot minimize $\cE_{X|Z}$ directly, since we do not observe samples from $\mu_{P_{X|Z}}$, but only the pairs $(x_i, z_i)$ from $(X,Z)$. We bound this with a surrogate loss $\Tilde{\cE}_{X|Z}$ that has a sample-based version:

\begin{equation*}
    \begin{split}
        \cE_{X|Z}(F) & = E_Z\left[||E_{X|Z}\left[k_{\cX}(X,\cdot) - F(Z)|Z\right]||_{\cH_\cX}^2\right] \\
        & \leq E_{Z}E_{X|Z}\left[||k_{\cX}(X,\cdot) - F(Z)||_{\cH_\cX}^2|Z\right] \\
        & = E_{X,Z}\left[||k_{\cX}(X,\cdot) - F(Z)||_{\cH_\cX}^2\right] \\ 
        & =: \Tilde{\cE}_{X|Z}(F)
    \end{split}
\end{equation*}

For details regarding the use of the surrogate loss function and its meaning, see \citet{park2020measure}. We empirically estimate the surrogate population loss $\Tilde{\cE}_{X|Z}$ using a regularized loss function $\Tilde{\cE}_{X|Z,N,\lambda}$  
for $\{(x^{(n)}, z^{(n)})\}_{n=1}^N$ from the joint distribution $P_{XZ}$,
\begin{equation}
    \Tilde{\cE}_{X|Z,N,\lambda}(F) \coloneqq \frac{1}{N} \sum_{n=1}^N ||k_{\cX}(x^{(n)},\cdot) - F(z^{(n)})||_{\cH_{\cX}}^2 + \lambda||F||_{\cG_{\cX\cZ}}^2 \; , 
\end{equation}
where $\lambda$ is a regularization parameter. We use the following theorem.

\begin{theorem}[Loss function \citep{micchelli2005learning}]\label{th2}
    Suppose we want to perform regression with input space $\cZ$ and output space $\cH$, by minimizing 
    \begin{equation*}
    \frac{1}{N} \sum_{n=1}^N ||h^{(n)} - F(z^{(n)})||_{\cH}^2 + \lambda||F||_{\cG}^2 
    \end{equation*}
    where $\lambda>0$ is a regularization parameter, $\cG$ is an $\cH$-valued RKHS on $\cZ$ with $\cH$-kernel $\Gamma$ and $\{(z^{(n)}, h^{(n)}): n = 1, \dots ,N \} \subseteq \cZ \times \cH$.
    \newline
    If $\Tilde{F}$ minimizes the above equation in $\cG$, it is unique and has the form $\Tilde{F} = \sum_{n=1}^N \Gamma(\cdot, z^{(n)})(u^{(n)})$ where the coefficients $\{u^{(n)}: n =1,\dots ,N\} \subseteq \cH$ are the unique solution of the linear equations $\sum_{n'=1}^N \left(\Gamma(z^{(n)},z^{(n')}) + N\lambda\delta_{n,n'}\right)(u^{(n')}) = h^{(n)}, n =1 ,\dots ,N$ ($\delta_{n,n'}$ is the Kronecker delta).
\end{theorem}
Our loss function matches the form in the Theorem~\ref{th2}. Therefore, by the Theorem~\ref{th2}, the minima $\hat{F}_{P_{X|Z},N,\lambda}$ of $\Tilde{\cE}_{X|Z,N,\lambda}$ is $\hat{F}_{P_{X|Z},N,\lambda}(\cdot) = \textbf{k}_{Z}^{T}(\cdot)\textbf{f}$ where $\textbf{k}_{Z}(\cdot) \coloneqq (k_{\cZ}(z^{(1)},\cdot),\dots, k_{\cZ}(z^{(N)},\cdot))^T $, $\textbf{f} \coloneqq (f^{(1)}, \dots, f^{(N)})^T$ and the coefficients $f^{(n)} \in \cH_{\cX}$ are the unique solutions of the linear equations $(\textbf{K}_Z+N\lambda\textbf{I})\textbf{f} = \textbf{k}_X$, where $\left[\textbf{K}_Z\right]_{ij} \coloneqq k_{\cZ}(z^{(i)},z^{(j)})$, $\textbf{k}_X \coloneqq (k_{\cX}(x^{(1)},\cdot), \dots, k_{\cX}(x^{(N)},\cdot))^T$ and $\textbf{I}$ is the $N\times N$ identity matrix. Hence, the coefficients are $\textbf{f} = \textbf{W}\textbf{k}_X$, where $\textbf{W} = (\textbf{K}_Z+N\lambda\textbf{I})^{-1}$. Finally, we get
\begin{equation*}
    \hat{F}_{P_{X|Z},N,\lambda}(\cdot) = \textbf{k}_{Z}^{T}(\cdot)\textbf{W}\textbf{k}_X \in \cG_{\cX\cZ}
\end{equation*}
We now construct the empirical estimator of the MCMD between the distributions $P_{X|Z}$ and $P_{X'|Z'}$. Given samples $\{(x^{(n)},z^{(n)})\}_{n=1}^N$,$\{(x'^{(n)},z'^{(n)})\}_{n=1}^N$ from distributions $P_{XZ},P_{X'Z'}$, we estimate the MCMD as
\begin{equation}
    \begin{split}
        \widehat{\text{MCMD}}_{P_{X|Z},P_{X'|Z'}}(\cdot) &= ||\hat{F}_{P_{X|Z},N,\lambda}(\cdot)-\hat{F}_{P_{X'|Z'},N,\lambda}(\cdot)||_{\cH_{\cX}}\\
        &=\left(\textbf{k}_{Z}^{T}(\cdot)\textbf{W}_Z\textbf{K}_X\textbf{W}_Z\textbf{k}_{Z}(\cdot) + \textbf{k}_{Z'}^{T}(\cdot)\textbf{W}_{Z'}\textbf{K}_{X'}\textbf{W}_{Z'}\textbf{k}_{Z'}(\cdot)\right.\\
        &\left.\quad- 2\textbf{k}_{Z}^{T}(\cdot)\textbf{W}_Z\textbf{K}_{XX'}\textbf{W}_{Z'}\textbf{k}_{Z'}(\cdot)\right)^{1/2}
    \end{split}
\end{equation}
where $[\textbf{K}_X]_{st}=k_{\cX}(x^{(s)},x^{(t)})$, $[\textbf{K}_{X'}]_{st}=k_{\cX}(x'^{(s)},x'^{(t)})$, $[\textbf{K}_{XX'}]_{st}=k_{\cX}(x^{(s)},x'^{(t)})$, $[\textbf{K}_{Z'}]_{st}=k_{\cZ}(z'^{(s)},z'^{(t)})$,
$\textbf{k}_{Z'}(\cdot)=(k_{\cZ}(z'^{(1)},\cdot),\dots,k_{\cZ}(z'^{(N)},\cdot))$, $\textbf{W}_{Z}=(\textbf{K}_Z+N\lambda\textbf{I})^{-1}$ and $\textbf{W}_{Z'}=(\textbf{K}_{Z'}+N\lambda\textbf{I})^{-1}$.

\section{Intervention mean embeddings}\label{IME}

\subsection{Definition}\label{3.1}
We derive the  mean embedding for the interventional distribution given in Equation~\eqref{eq: 1}. Recall that $X_d: \Omega \rightarrow \cX_d,\, 1\leq d \leq D$ are random variables where $(\cX_d,\mathfrak{X}_d)$ are separable measurable spaces. For $1\leq d \leq D$, $\cH_{\cX_d}$ denotes the RKHS of functions on $\cX_d$ with reproducing kernel $k_{\cX_d}(\cdot,\cdot)$.  For an intervened node $X_i$, target node $X_j$ and a valid adjustment set $\mathbf{Z}$ for the pair $(X_i,X_j)$, $j \neq i$, let $\mu_{P_{X_j|do(X_i)=\hat{x}_i}}$ denote the intervention mean embedding (IME) corresponding to the interventional distribution $P_{X_j|do(X_i)=\hat{x}_i}$.
Let $\mu_{P_{X_j|X_i,\textbf{Z}}} = \mathbb{E}_{X_j|X_i,\textbf{Z}}[k_{\cX_j}(X_j,\cdot)|X_i,\textbf{Z}]$. Then, by Theorem \ref{th1}, we can write $\mu_{P_{X_j|X_i,\textbf{Z}}} = F_{P_{X_j|X_i,\textbf{Z}}} \circ (X_i,\textbf{Z})$, where $F_{P_{X_j|X_i,\textbf{Z}}}: \cX_i \times \boldsymbol{\cZ} \rightarrow \cH_{\cX_j}$ is some deterministic function measurable with respect to $\mathfrak{X}_i \times \boldsymbol{\mathfrak{Z}}$ and $\mathcal{B}(\cH_{\cX_j})$. 
\begin{align}
    \mu_{P_{X_j|do(X_i)=\hat{x}_i}} &\coloneqq \int_{\cX_j} k_{\cX_j}(x_j,\cdot) p(x_j|do(X_i)=\hat{x}_i) dx_j \label{eq3} \\
    &= \int_{\cX_j} k_{\cX_j}(x_j,\cdot) \left( \int_{\boldsymbol{\cZ}}p(x_j|\hat{x}_i,\textbf{z})p(\textbf{z})d\textbf{z} \right) dx_j \label{eq4} \\
    &= \int_{\boldsymbol{\cZ}} \left( \int_{\cX_j} k_{\cX_j}(x_j,\cdot)p(x_j|\hat{x}_i,\textbf{z})dx_j \right) p(\textbf{z})d\textbf{z} \label{eq5} \\
    &= \int_{\boldsymbol{\cZ}} F_{P_{X_j|X_i,\textbf{Z}}}(\hat{x}_i,\textbf{z}) p(\textbf{z}) d\textbf{z} \label{eq6} \\
    &= \mathbb{E}_{\textbf{Z}}\left[F_{P_{X_j|X_i,\textbf{Z}}}(\hat{x}_i,\textbf{Z})\right] 
\end{align}
\par
Equation~\eqref{eq3} follows from the definition of mean embedding of a distribution in Equation~\eqref{eq2}, Equation~\eqref{eq4} follows from the expression for interventional distribution in Equation~\eqref{eq: valid_adjust}, Equation~\eqref{eq5} involves interchanging the order of integration and Equation~\eqref{eq6} follows from Theorem~\ref{th1}. 

Let $G_{P_{X_j|do(X_i)}}(\cdot) = \mathbb{E}_{\textbf{Z}}[F_{P_{X_j|X_i,\textbf{Z}}}(X_i,\textbf{Z})]$, then $G_{P_{X_j|do(X_i)}}: \cX_i \rightarrow \cH_{\cX_j}$ is a measurable, deterministic function and maps each possible intervention $\hat{x}_i \in \cX_i$ to the embedding of its interventional distribution $P_{X_j|do(X_i)=\hat{x}_i}$, i.e., it is the family of embeddings of interventional distributions. Let $P_{X_j|do(X_i)}$ and $P'_{X_j|do(X_i)}$ be the interventional distributions for two different valid adjustment sets (as is the case when we consider the distribution of $X_j$ after intervening on $X_i$ in two different DAGs). The MCMD between these distributions is $\text{MCMD}_{P_{X_j|do(X_i)},P'_{X_j|do(X_i)}}(\cdot) = ||G_{P_{X_j|do(X_i)}}(\cdot)-G_{P'_{X_j|do(X_i)}}(\cdot)||_{\cH_{\cX_j}}$ where $\text{MCMD}_{P_{X_j|do(X_i)},P'_{X_j|do(X_i)}}(\cdot):\cX_i\to\mathbb{R}$.

\subsection{Empirical estimate} \label{empirical_estimate}

First we compute the empirical estimate for $F_{P_{X_j|X_i,\textbf{Z}}}$. This follows based on the derivation in section \ref{empirical-estimates} where instead of conditioning only on one variable, we condition on $X_i$ and $\textbf{Z}$. We aim to find the minima of the loss function $\cE_{X_j|X_i,\textbf{Z}}(F) = \mathbb{E}_{X_i,\textbf{Z}}\left[||F(X_i,\textbf{Z}) - F_{P_{X_j|X_i,\textbf{Z}}}(X_i,\textbf{Z})||^{2}_{\cH_{\cX_j}}\right]$ among all $F \in \cG_{\cX_j,\cX_i\boldsymbol{\cZ}}$ where $\cG_{\cX_j,\cX_i\boldsymbol{\cZ}}$ is the RKHS of functions from $\cX_i \times \boldsymbol{\cZ}$ to $\cH_{\cX_j}$. We endow $\cG_{\cX_j,\cX_i\boldsymbol{\cZ}}$ with the kernel $l_{\cX_j,\cX_i\boldsymbol{\cZ}}((x_i,\textbf{z}),(x'_i,\textbf{z}'))=k_{\cX_i\boldsymbol{\cZ}}((x_i,\textbf{z}),(x'_i,\textbf{z}'))\textbf{Id}$ where $k_{\cX_i\boldsymbol{\cZ}}$ is a kernel on $\cX_i\times\boldsymbol{\cZ}$ (see Remark \ref{rem: 1}).
\begin{equation*}
    \begin{split}
        \cE_{X_j|X_i,\textbf{Z}}(F) &= \mathbb{E}_{X_i,\textbf{Z}}\left[||\mathbb{E}_{X_j|X_i,\textbf{Z}}\left[k_{\cX_j}(X_j,\cdot) - F(X_i,\textbf{Z})\right]|X_i,\textbf{Z}||^{2}_{\cH_{\cX_j}}\right] \\ 
        &\leq \mathbb{E}_{X_i,\textbf{Z}}\mathbb{E}_{X_j|X_i,\textbf{Z}}\left[||k_{\cX_j}(X_j, \cdot) - F(X_i,\textbf{Z})||^{2}_{\cH_{\cX_j}}|X_i,\textbf{Z}\right] \\ 
        &= \mathbb{E}_{X_i,X_j,\textbf{Z}}\left[||k_{\cX_j}(X_j,\cdot) - F(X_i,\textbf{Z})||^{2}_{\cH_{\cX_j}}\right] \\
        &=: \Tilde{\cE}_{X_j|X_i,\textbf{Z}}(F) \\   
    \end{split}      
\end{equation*}
Since we do not observe samples from $\mu_{P_{X_j|X_i,\textbf{Z}}}$, instead of directly finding the minima of $\cE_{X_j|X_i,\textbf{Z}}$, we solve for the minima of the surrogate loss function $\Tilde{\cE}_{X_j|X_i,\textbf{Z}}$. The empirical regularized version of the surrogate loss function is given by $\hat{\cE}_{X_j|X_i,\textbf{Z},N,\lambda}(F) \coloneqq \frac{1}{N} \sum_{n=1}^N ||k_{\cX_j}(x_j^{(n)}, \cdot) - F(x_i^{(n)},\textbf{z}^{(n)})||^{2}_{\cH_{\cX_j}} + \lambda ||F||^{2}_{\cG_{\cX_j,\cX_i\boldsymbol{\cZ}}}$ where $\{x_i^{(n)},x_j^{(n)},\textbf{z}^{(n)}\}_{n=1}^N$ are samples from the joint distribution $P_{X_iX_j\textbf{Z}}$.
From Theorem \ref{th2},  the minima $\Hat{F}_{P_{X_j|X_i,\textbf{Z},N,\lambda}}$ of $\hat{\cE}_{X_j|X_i,\textbf{Z},N,\lambda}$ is $\Hat{F}_{P_{X_j|X_i,\textbf{Z},N,\lambda}}(\cdot,\cdot) = \textbf{k}_{X_i\textbf{Z}}^{T}(\cdot,\cdot)\textbf{f}$ where 
\begin{align}\label{eq : prod_k}
    \textbf{k}_{X_i\textbf{Z}}(\cdot,\cdot) &\coloneqq (k_{\cX_i\boldsymbol{\cZ}}((x_i^{(1)},\textbf{z}^{(1)}),(\cdot,\cdot)), \dots, k_{\cX_i\boldsymbol{\cZ}}((x_i^{(N)},\textbf{z}^{(N)}),(\cdot,\cdot)))^T 
\end{align} 
$\textbf{f} \coloneqq (f^{(1)}, \dots, f^{(N)})^T$ and $f^{(i)} \in \cH_{\cX_j}$ are unique solutions of the linear equation
\begin{align*}
    (\textbf{K}_{X_iZ} + N\lambda\textbf{I})\textbf{f} = \textbf{k}_{X_j}
\end{align*}
where $[\textbf{K}_{X_i\textbf{Z}}]_{st} \coloneqq k_{\cX_i\boldsymbol{\cZ}}((x_i^{(s)},\textbf{z}^{(s)}),(x^{(t)},\textbf{z}^{(t)}))$ and $\textbf{k}_{X_j} \coloneqq (k_{\cX_j}(x_j^{(1)},\cdot), \dots, k_{\cX_j}(x_j^{(N)},\cdot))^T$. Hence $\textbf{f} = \textbf{W}\textbf{k}_{X_j}$ where $\textbf{W} = (\textbf{K}_{X_i\textbf{Z}}+N\lambda \textbf{I})^{-1}$.
Therefore, $\Hat{F}_{P_{X_j|X_i,\textbf{Z},N,\lambda}}(\cdot,\cdot) = \textbf{k}_{X_i\textbf{Z}}(\cdot,\cdot)\textbf{W}\textbf{k}_{X_j}$.
\par
Using $\Hat{F}_{P_{X_j|X_i,\textbf{Z},N,\lambda}}$, we obtain the empirical estimate for $G_{P_{X_j|do(X_i)}}: \cX_i \rightarrow \cH_{\cX_j}$.
\begin{align*}
    \hat{G}_{P_{X_j|do(X_i)}}(\cdot) = \frac{1}{N}\sum_{n=1}^N \textbf{k}_{X_i\textbf{Z}}^T(\cdot,\textbf{z}^{(n)})\textbf{W}\textbf{k}_{X_j}
\end{align*}
If $P_{X_j|do(X_i)}$ and $P'_{X_j|do(X_i)}$ are the interventional distributions for two different valid adjustment sets $Z$ and $Z'$, their MCMD  can be computed as follows: given samples $\{(x^{(n)}_i,x^{(n)}_j,z^{(n)})\}_{n=1}^N$ and $\{(x^{(n)}_i,x^{(n)}_j,z'^{(n)})\}_{n=1}^N$ from $P_{X_iX_jZ}$ and $P_{X_iX_jZ'}$, the MCMD can be estimated as:
\begin{equation}
    \begin{split}
        \widehat{MCMD}_{P_{X_j|do(X_i)},P'_{X_j|do(X_i)}}(\cdot) &= ||\hat{G}_{P_{X_j|do(X_i)}}(\cdot)-\hat{G}_{P'_{X_j|do(X_i)}}(\cdot)||_{\cH_{\cX_j}}\\
        &=\left[\left(\frac{1}{N}\sum_{n=1}^N \textbf{k}_{X_i\textbf{Z}}^T(\cdot,\textbf{z}^{(n)})\right)\textbf{W}_{\textbf{Z}}\textbf{K}_{X_j}\textbf{W}_{\textbf{Z}}\left(\frac{1}{N}\sum_{n=1}^N \textbf{k}_{X_i\textbf{Z}}(\cdot,\textbf{z}^{(n)})\right)\right.\\
        &\left.\quad+\left(\frac{1}{N}\sum_{n=1}^N \textbf{k}_{X_i\textbf{Z}'}^T(\cdot,\textbf{z}'^{(n)})\right)\textbf{W}_{\textbf{Z}'}\textbf{K}_{X_j}\textbf{W}_{\textbf{Z}'}\left(\frac{1}{N}\sum_{n=1}^N \textbf{k}_{X_i\textbf{Z}'}(\cdot,\textbf{z}'^{(n)})\right)\right.\\
        &\left.\quad-2\left(\frac{1}{N}\sum_{n=1}^N \textbf{k}_{X_i\textbf{Z}}^T(\cdot,\textbf{z}^{(n)})\right)\textbf{W}_{\textbf{Z}}\textbf{K}_{X_j}\textbf{W}_{\textbf{Z}'}\left(\frac{1}{N}\sum_{n=1}^N \textbf{k}_{X_i\textbf{Z}'}(\cdot,\textbf{z}'^{(n)})\right)\right]^{1/2}\\
    \end{split}
\end{equation}
where $[\textbf{K}_{X_j}]_{st}=k_{\cX_j}(x_j^{(s)},x_j^{(t)})$, $\textbf{W}_{\textbf{Z}}=(\textbf{K}_{X_i\textbf{Z}}+N\lambda\textbf{I})^{-1}$, $\textbf{W}_{\textbf{Z}'}=(\textbf{K}_{X_i\textbf{Z}'}+N\lambda\textbf{I})^{-1}$, $[\textbf{K}_{X_i\textbf{Z}'}]_{st} \coloneqq k_{\cX_i\boldsymbol{\cZ}}((x_i^{(s)},\textbf{z}'^{(s)}),(x^{(t)},\textbf{z}'^{(t)}))$ and $\textbf{k}_{X_i\textbf{Z}'}(\cdot,\cdot) \coloneqq (k_{\cX_i\boldsymbol{\cZ}}((x_i^{(1)},\textbf{z}'^{(1)}),(\cdot,\cdot)), \dots,\\ k_{\cX_i\boldsymbol{\cZ}}((x_i^{(N)},\textbf{z}'^{(N)}),(\cdot,\cdot)))^T $
\begin{remark}[Product kernels]\label{rem: 1}
     We can choose $k_{\cX_i\boldsymbol{\cZ}}$ to be the product kernel:
     \begin{equation}
         k_{\cX_i\boldsymbol{\cZ}}((x_i,\textbf{z}),(x'_i,\textbf{z}')) = k_{\cX_i}(x_i,x'_i)k_{\boldsymbol{\cZ}}(\textbf{z},\textbf{z}')
     \end{equation}
      Let $|Z| = M \text{ so that }\textbf{Z} = \{X_{i_1},\dots,X_{i_M}\}$.  
      Given reproducing kernels $k_{\cX_d}$ of RKHSs $\cH_{\cX_d}$, $1\leq d\leq D$, we can also choose $k_{\boldsymbol{\cZ}}$ to be the product kernel:
      \begin{equation}
          k_{\boldsymbol{\cZ}}(\textbf{z},\textbf{z}') = k_{\cX_{i_1}}(x_{i_1},x'_{i_1})\dots k_{\cX_{i_M}}(x_{i_M},x'_{i_M})
      \end{equation}
\end{remark}

\section{Continuous structural intervention distance} \label{contSID}

Consider the setting where we have a true DAG $\cG_1 = (\mathbf{V},\cE_{\cG_1})$, a learnt DAG $\cG_2 = (\mathbf{V},\cE_{\cG_2})$ and observational data $\mathcal{D}$ sampled from an unknown distribution $P$ with density $p(\cdot)$ that is Markov with respect to $\cG_1$ and $\cG_2$ (see Definition \ref{def1}). 
Note that the true and learnt DAGs have a common set of vertices but differ in their edges. 
Let $P_{X_j|do(X_i);\cG_1}$ and $P_{X_j|do(X_i);\cG_2}$ denote the interventional distribution corresponding to intervening on $X_i$ and observing $X_j$ in the true DAG $\cG_1$ and the learnt DAG $\cG_2$, respectively. 
The densities of both these distributions can be calculated from $p(\cdot)$ using the adjustment formula \eqref{eq: valid_adjust} and taking $\mathbf{Z}$ to be $\mathbf{PA}_i$, the parent set of $X_i$.

First, we generate the set $\mathbf{V}^2 \coloneqq (\mathbf{V} \times \mathbf{V})$, which consists of all ordered pairs of nodes from the common vertex set of the true DAG and the learnt DAG. 
For each pair $(X_i,X_j) \in \mathbf{V}^2,i\neq j$, we compare the distribution of $X_j$ obtained by intervening on $X_i$ in $\cG_1$ and $\cG_2$ (this can be extended to multiple simultaneous interventions---see Remark \ref{rem: mult_int}). 
Unless otherwise stated, we use the observational data of $X_i$ as our interventions while comparing the interventional distributions between the true DAG and the learnt DAG (one may specify a different distribution on the interventions---see Remark \ref{rem: prior_dist}). 
We record the difference in a function $d:\Tilde{\mathbf{V}}^2 \rightarrow \mathbb{R}_{\geq 0}$ which we describe below by examining various possible cases.

\textit{Case 1: }There is no directed path from $X_i$ to $X_j$ in DAGs $\cG_1$ and $\cG_2$ (in Algorithm~\ref{alg:distance} denoted as ``$\text{checkDirectedPath}(X_i,X_j,\cG$)''). 
In the absence of a directed path from the intervened node to the target node, an intervention has no effect on the target node. 
So, in $\cG_1$ and $\cG_2$ the distribution of $X_j$ obtained by intervening on $X_i$ is equal to the observational distribution of $X_j$, i.e., $P_{X_j|do(X_i);\cG_1} = P_{X_j|do(X_i);\cG_2} = P_{X_j}$. This in turn implies $d(X_i,X_j) = 0$.

\textit{Case 2: }There is a directed path from $X_i$ to $X_j$ in $\cG_1$ but not in $\cG_2$. 
The same argument used in Case 1 can be applied here to obtain $P_{X_j|do(X_i);\cG_2}=P_{X_j}$. 
Intervening on $X_i$ has an effect on $X_j$ in $\cG_1$ due to the presence of the directed path $X_i \rightarrow X_j$ and the resulting distribution can be computed by adjusting for the parent set of $X_i$ in $\cG_1$, i.e., $\textbf{PA}_{i,\cG_1}$. 
We compare the two distributions $P_{X_j|do(X_i);\cG_1}$ and $P_{X_j}$ by computing the average over their MMDs for each observed $x_i$. 
We then divide by the norm of the embedding of the observational distribution $X_j$ to make contSID scale-invariant. 
The resulting distance $d$ is defined as we state in Equation~\eqref{eq : case2}, where we denote $\sum_{m,m'=1}^N k_{\cX_j}(x^{(m)}_j,x^{(m')}_j)$ by $C_{X_j}$.

\begin{equation}\label{eq : case2}
    \begin{split}
        d(X_i,X_j) &= \frac{1}{N}\sum_{n=1}^N ||\Tilde{\mu}_{P_{X_j|do(X_i)=x^{(n)}_i;\cG_1}}-\Tilde{\mu}_{P_{X_j}}||_{\cH_{\cX_j}}\\
        &= \frac{1}{N}\sum_{n=1}^N||\frac{1}{N}\sum_{m=1}^N \textbf{k}^T_{X_i\textbf{PA}_{i,\cG_1}}(x^{(n)}_i,\textbf{pa}^{(m)}_{i,\cG_1})\textbf{W}_{\cG_1}\textbf{k}_{X_j}(\cdot)-\frac{1}{N}\sum_{m'=1}^N k_{\cX_j}(x^{(m')}_j,\cdot)||_{\cH_{\cX_j}}\\
        &= \frac{1}{N\sqrt{C_{X_j}}}\sum_{n=1}^N\left[ \left(\sum_{m=1}^N \textbf{k}^T_{X_i\textbf{PA}_{i,\cG_1}}(x^{(n)}_i,\textbf{pa}^{(m)}_{i,\cG_1})\right)\textbf{W}_{\cG_1}\textbf{K}_{X_j}\textbf{W}_{\cG_1}\left(\sum_{m=1}^N \textbf{k}^T_{X_i\textbf{PA}_{i,\cG_1}}(x^{(n)}_i,\textbf{pa}^{(m)}_{i,\cG_1})\right)\right.\\
        &\left. + C_{X_j} -2\left(\sum_{m=1}^N \textbf{k}^T_{X_i\textbf{PA}_{i,\cG_1}}(x^{(n)}_i,\textbf{pa}^{(m)}_{i,\cG_1})\right)\textbf{W}_{\cG_1}\left(\sum_{m=1}^N\textbf{k}_{X_j}(x^{(m)}_j)\right)\right]^{1/2}
    \end{split}
\end{equation}
Similarly, if there is a directed path from $X_i$ to $X_j$ in $\cG_2$ but not in $\cG_1$, the resulting distance $d$ is:
\begin{equation}\label{eq : case3}
    \begin{split}
        d(X_i,X_j) &= \frac{1}{N\sqrt{C_{X_j}}}\sum_{n=1}^N\left[ \left(\sum_{m=1}^N \textbf{k}^T_{X_i\textbf{PA}_{i,\cG_2}}(x^{(n)}_i,\textbf{pa}^{(m)}_{i,\cG_2})\right)\textbf{W}_{\cG_2}\textbf{K}_{X_j}\textbf{W}_{\cG_2}\left(\sum_{m=1}^N \textbf{k}^T_{X_i\textbf{PA}_{i,\cG_2}}(x^{(n)}_i,\textbf{pa}^{(m)}_{i,\cG_2})\right)\right.\\
        &\left.+ C_{X_j}-2\left(\sum_{m=1}^N \textbf{k}^T_{X_i\textbf{PA}_{i,\cG_2}}(x^{(n)}_i,\textbf{pa}^{(m)}_{i,\cG_2})\right)\textbf{W}_{\cG_2}\left(\sum_{m=1}^N\textbf{k}_{X_j}(x^{(m)}_j)\right)\right]^{1/2}       
    \end{split}
\end{equation}

\textit{Case 3: }There is a directed path from $X_i$ to $X_j$ in DAG $\cG_1$ and $\cG_2$. 
The distribution of $X_j$ after intervening on $X_i$ in $\cG_1$ can be computed by adjusting for the parent set of $X_i$ in $\cG_1$ - $\textbf{PA}_{i;\cG_1}$. Similarly, we obtain the interventional distribution of $X_j$ in $\cG_2$ by adjusting for the parent set of $X_i$ in $\cG_2$ - $\textbf{PA}_{i;\cG_2}$. 
\begin{enumerate}
    \item If $\textbf{PA}_{i;\cG_1}$ is a valid adjustment set (Definition~\ref{def0}) in $\cG_2$ or $\textbf{PA}_{i;\cG_2}$ is a valid adjustment set in $\cG_1$, then by \eqref{eq: valid_adjust}, $P_{X_j|do(X_i);\cG_1} = P_{X_j|do(X_i);\cG_2}$, hence $d(X_i,X_j)=0$.\footnote{In general, the above condition is not necessary for $P_{X_j|do(X_i);\cG_1} = P_{X_j|do(X_i);\cG_2}$. 
    It is sufficient that there is a common valid adjustment set---not just a parent adjustment set---for the pair $(X_i,X_j)$ in $\cG_1$ and $\cG_2$. 
    However, it is not straightforward and beyond the scope of this article to compare the validity of an adjustment in different DAGs. 
    Thus, we resort to the simple and inexpensive graphical task of checking if the parent sets in one DAG are valid adjustment sets in the other DAG.}
    
    \item If $\textbf{PA}_{i;\cG_1}$ is not a valid adjustment set in $\cG_2$ or $\textbf{PA}_{i;\cG_2}$ is not a valid adjustment set in $\cG_1$, then the interventional distributions $P_{X_j|do(X_i);\cG_1}$ and $P_{X_j|do(X_i);\cG_2}$ \emph{may not} be equal. 
    To assess the difference, we compute the average over their MMDs for each $x_i \sim \cD_i$. 
    We divide by the norm of the embedding of the observational distribution $X_j$ to make contSID scale-invariant. The resulting distance $d$ is defined as we state in Equation~\eqref{eq : case4}.

    \begin{equation}\label{eq : case4}
        \begin{split}
            d(X_i,X_j) &= \frac{1}{N}\sum_{n=1}^N ||\Tilde{\mu}_{P_{X_j|do(X_i=x^{(n)}_i);\cG_2}}-\Tilde{\mu}_{P_{X_j|do(X_i=x^{(n)}_i);\cG_1}}||_{\cH_{\cX_j}}\\
            &= \frac{1}{N}\sum_{n=1}^N||\frac{1}{N}\sum_{m=1}^N \textbf{k}^T_{X_i\textbf{PA}_{i,\cG_2}}(x^{(n)}_i,\textbf{pa}^{(m)}_{i,\cG_2})W_{\cG_2}\textbf{k}_{X_j}\\
            &\qquad\qquad\qquad -\frac{1}{N}\sum_{m'=1}^N \textbf{k}^T_{X_i\textbf{PA}_{i,\cG_1}}(x^{(n)}_i,\textbf{pa}^{(m')}_{i,\cG_1})W_{\cG_1}\textbf{k}_{X_j} ||_{\cH_{\cX_j}}\\
            &= \frac{1}{N^2}\sum_{n=1}^N\left[ \left(\sum_{m=1}^N \textbf{k}^T_{X_i\textbf{PA}_{i,\cG_2}}(x^{(n)}_i,\textbf{pa}^{(m)}_{i,\cG_2})\right)\textbf{W}_{\cG_2}\textbf{K}_{X_j}\textbf{W}_{\cG_2}\left(\sum_{m=1}^N \textbf{k}^T_{X_i\textbf{PA}_{i,\cG_2}}(x^{(n)}_i,\textbf{pa}^{(m)}_{i,\cG_2})\right)\right.\\
            & +\left(\sum_{m=1}^N \textbf{k}^T_{X_i\textbf{PA}_{i,\cG_1}}(x^{(n)}_i,\textbf{pa}^{(m)}_{i,\cG_1})\right)\textbf{W}_{\cG_1}\textbf{K}_{X_j}\textbf{W}_{\cG_1}\left(\sum_{m=1}^N \textbf{k}^T_{X_i\textbf{PA}_{i,\cG_1}}(x^{(n)}_i,\textbf{pa}^{(m)}_{i,\cG_1})\right)\\
            &\left.-2\left(\sum_{m=1}^N \textbf{k}^T_{X_i\textbf{PA}_{i,\cG_2}}(x^{(n)}_i,\textbf{pa}^{(m)}_{i,\cG_2})\right)\textbf{W}_{\cG_2}\textbf{K}_{X_j}\textbf{W}_{\cG_1}\left(\sum_{m=1}^N \textbf{k}^T_{X_i\textbf{PA}_{i,\cG_1}}(x^{(n)}_i,\textbf{pa}^{(m)}_{i,\cG_1})\right)\right]^{1/2}
        \end{split}
    \end{equation}

\end{enumerate}

We summarise the various cases and the applicable equations in Algorithm~\ref{alg:distance}.
In Algorithm~\ref{alg:contSID}, we describe that the contSID is calculated over each ordered pair $(X_i, X_j) \in \textbf{V}^2, i \neq j$.

\begin{algorithm}
\caption{$d(X_i,X_j,\cG_1,\cG_2,\cD)$} \label{alg:distance}
\begin{algorithmic}[1]
\Input Intervened node $X_i$, target node $X_j$, true DAG $\cG_1 = (\mathbf{V},E_{\cG_1})$, learnt DAG $\cG_2 = (\mathbf{V},E_{\cG_2})$ and the observational data $\mathcal{D}$
\State $c_{\cG_1} \gets \text{checkDirectedPath}(X_i,X_j,\cG_1)$
\State $c_{\cG_2} \gets \text{checkDirectedPath}(X_i,X_j,\cG_2)$
\If{$c_{\cG_1}==\text{False and }c_{\cG_2}==\text{False}$}
    \State \Return 0
\Else
    \State $Z_{\cG_1} \gets \textbf{PA}_{i,\cG_1}$
    \State $Z_{\cG_2} \gets \textbf{PA}_{i,\cG_2}$
    \State $K \gets \sum_{m,m'}^N k(x^{(m)}_j,x^{(m')}_j)$
    \If{$c_{\cG_1}==\text{True and }c_{\cG_2}==\text{False}$}
        \State \Return \eqref{eq : case2}
    \ElsIf{$c_{\cG_1}==\text{False and }c_{\cG_2}==\text{True}$}
        \State \Return \eqref{eq : case3}
    \Else
        \If{$Z_{\cG_1}$ is a valid adjustment set in $\cG_2$ or $Z_{\cG_2}$ is a valid adjustment set in $\cG_1$}
            \State \Return 0
        \Else
            \State \Return \eqref{eq : case4}
        \EndIf
    \EndIf
\EndIf

\end{algorithmic}
\end{algorithm}

\begin{algorithm}
\caption{$\text{contSID}(\cG_1,\cG_2,\cD)$} \label{alg:contSID}
\begin{algorithmic}[1]
\Input True DAG $\cG_1 = (\mathbf{V},E_{\cG_1})$, learnt DAG $\cG_2 = (\mathbf{V},E_{\cG_2})$ and the observational data $\mathcal{D}$
\State $\text{sum} \gets 0$
\For{$(X_i,X_j) \in \mathbf{V}^2,\quad i\neq j$}
    \State $\text{sum} = \text{sum} + d(X_i,X_j,\cG_1,\cG_2,\cD)$
\EndFor
\State \Return sum
\end{algorithmic}
\end{algorithm}

\begin{remark}[Interventions on multiple variables]\label{rem: mult_int}
As in \citet{peters2015structural}, we have considered intervening on single variables only. 
However, the contSID can be extended to account for interventions on multiple variables as well. 
Since the union of parent sets of the intervened variables is not necessarily a valid adjustment set, one would need to define a valid adjustment set for the intervened variables and the observed variable. 
Then, using a modified version of Equation~\eqref{eq: valid_adjust}, we can compute the interventional distribution and its corresponding embedding. 
This can be achieved by replacing the one intervened variable $X_i:\Omega\to\cX$ with the set of variables $\textbf{X}_i:\Omega\to\boldsymbol{\cX}_i$ that we intervene on, and defining the corresponding kernel $k_{\boldsymbol{\cX}_i}:{\boldsymbol{\cX}_i}\times{\boldsymbol{\cX}_i}\to\mathbb{R}$.
\end{remark}
\begin{remark}[Prior distribution on interventions]\label{rem: prior_dist}
Unless specified, the computation of the contSID uses the empirical distribution of $X_i$ to compute the average of the MMDs in Equations~\eqref{eq : case2}, \eqref{eq : case3} and \eqref{eq : case4}. 
If required, however, one may specify an alternative distribution on the intervention, e.g., assigning measure 1 to a single intervention, and evaluate the contSID with that interventional distribution.
\end{remark}

\section{Experiments} \label{experiments}

For each number of nodes $p \in\{5,10,20\}$, we generate 100 DAGs by an Erdos-R\`enyi model with the probability of the existence of an edge equal to 0.25. 
100 \emph{iid} samples $\mathcal{D} \in \mathbb{R}^p$ are generated for each DAG according to a linear SEM with non-Gaussian (exponential) noise. Linear coefficients are sampled uniformly from the interval $[-10,10]$ and the exponential noise has scale $\beta=1$. For each simulated DAG, we obtain predicted DAGs by running the PC (constraint-based), GES (score-based) and ICALiNGAM (function-based causal discovery algorithms) \citep[respectively]{spirtes2000causation, chickering2002optimal, shimizu2006linear} on the synthetically generated data. 
We compute the average SHD, SID and contSID values as well as their standard deviation for each true and learnt DAG pair. The ICALiNGAM algorithm outperforms PC and GES algorithms across all nodes and all metrics (SHD, SID and contSID). However, while both SHD and SID indicate that the GES algorithm outperforms the PC algorithm (for $p=10,20$), contSID suggests the opposite, namely, that the PC algorithm is more accurate than the GES algorithm.
\begin{table}[h!]
    \centering
    \begin{tabular}{l l l l}
        \hline
        $p$ & PC & GES & ICALiNGAM \\
        \hline
        5 & 2.13 $\pm$ 1.32 & 2.18 $\pm$ 1.51 & 0.89 $\pm$ 1.04 \\
        10 & 10.29 $\pm$ 3.77 & 9.67 $\pm$ 4.88 & 3.55 $\pm$ 3.34 \\
        20 & 53.1 $\pm$ 7.12 & 47.6 $\pm$ 8.87 & 31.15 $\pm$ 10.65 \\
        \hline
    \end{tabular}
    \caption{Average SHD to true DAG for 100 simulations, for different values of $p$}
    \label{tab:SHD_table}
\end{table}
\begin{table}[h!]
    \centering
    \begin{tabular}{l l l l}
        \hline
        $p$ & PC & GES & ICALiNGAM \\
        \hline
        5 & 4.7 $\pm$ 3.76 & 4.45 $\pm$ 3.82 & 1.4 $\pm$ 2.20 \\
        10 & 37.21 $\pm$ 17.65 & 25.87 $\pm$ 14.60 & 7.86 $\pm$ 7.65 \\
        20 & 267.85 $\pm$ 39.02 & 248.23 $\pm$ 34.05 & 124.7 $\pm$ 41.50 \\
        \hline
    \end{tabular}
    \caption{Average SID to true DAG for 100 simulations, for different values of $p$}
    \label{tab:SID_table}
\end{table}
\begin{table}[h!]
    \centering
    \begin{tabular}{l l l l}
        \hline
        $p$ & PC & GES & ICALiNGAM \\
        \hline
        5 & 2.43 $\pm$ 1.98 & 2.51 $\pm$ 2.11 & 0.48 $\pm$ 0.63\\
        10 & 20.18 $\pm$ 9.35 & 23.45 $\pm$ 12.49 & 5.28 $\pm$ 5.40 \\
        20 & 83.30 $\pm$ 37.12 & 134.37 $\pm$ 41.89 & 51.04 $\pm$ 21.60 \\
        \hline
    \end{tabular}
    \caption{Average contSID to true DAG for 100 simulations, for different values of $p$}
    \label{tab:contSID_table}
\end{table}

\section{Conclusion} \label{Conclusion}
We propose a novel metric to accurately compare a learnt to a true directed acyclic graph (DAG) in causal structure learning settings. 
Albeit the widespread use of the structural Hemming distance (SHD) and the structural intervention distance (SID), two metrics that fulfil the purpose of comparing a learnt to a true DAG, they are based on graph properties only. 
Besides graph properties, our metric takes additionally the underlying data of the causal system into account and can, hence, distinguish between the importance of learning edges more accurately. 
The metric is defined as a distance between kernel conditional mean embeddings that are derived through a measure-theoretic approach.
We hope that researchers working on causal structure learning problems find our novel metric useful in their assessment of the accuracy of causal discovery algorithms, and that it can provide additional insights beyond the capabilities of the SHD and SID.

\clearpage

\clearpage
\bibliography{bib/bibliography}

\end{document}